\documentclass[10pt,journal,compsoc]{IEEEtran}
\usepackage{amsmath,amsfonts}
\usepackage{algorithmic}
\usepackage{algorithm}
\usepackage{multirow}
\usepackage{ragged2e}
\usepackage{array}
\usepackage[caption=false,font=normalsize,labelfont=sf,textfont=sf]{subfig}
\usepackage{textcomp}
\usepackage{stfloats}
\usepackage{url}
\usepackage{verbatim}
\usepackage{graphicx}
\usepackage{hyperref}
\usepackage{cite}
\usepackage[switch]{lineno}
\begin{document}

\title{Context Perception Parallel Decoder \\ for Scene Text Recognition}


\author{Yongkun~Du,
        Zhineng~Chen,\IEEEmembership{~Member,~IEEE,}
        Caiyan~Jia,
        Xiaoting~Yin,
        Chenxia~Li,
        Yuning~Du,
        and~Yu-Gang~Jiang,~\IEEEmembership{Senior~Member,~IEEE}

\IEEEcompsocitemizethanks{
\IEEEcompsocthanksitem{Yongkun Du, Zhineng Chen and Yu-Gang Jiang are with the School of Computer Science, Fudan University, Shanghai 200433, China.
E-mail: ykdu23@m.fudan.edu.cn, \{zhinchen, ygj\}@fudan.edu.cn.}
\IEEEcompsocthanksitem{Caiyan Jia is with the School of Computer and Information Technology, Beijing Jiaotong University, Beijing 100044, China.
E-mail: cyjia@bjtu.edu.cn.}
\IEEEcompsocthanksitem{Xiaoting Yin, Chenxia Li and Yuning Du are with Baidu Inc., Beijing 100085, China.
E-mail:\{yinxiaoting, lichenxia, duyuning\}@baidu.com.}
}
\thanks{Corresponding author: Zhineng~Chen.}
}


\markboth{For review only}%
{Shell \MakeLowercase{\textit{et al.}}: A Sample Article Using IEEEtran.cls for IEEE Journals}


\IEEEtitleabstractindextext{%
\begin{abstract}

Scene text recognition (STR) methods have struggled to attain high accuracy and fast inference speed. Autoregressive (AR)-based models implement the recognition in a character-by-character manner, showing superiority in accuracy but with slow inference speed. Alternatively, parallel decoding (PD)-based models infer all characters in a single decoding pass, offering faster inference speed but generally worse accuracy. We first present an empirical study of AR decoding in STR, and discover that the AR decoder not only models linguistic context, but also provides guidance on visual context perception. Consequently, we propose Context Perception Parallel Decoder (CPPD) to predict the character sequence in a PD pass. CPPD devises a character counting module to infer the occurrence count of each character, and a character ordering module to deduce the content-free reading order and placeholders. Meanwhile, the character prediction task associates the placeholders with characters. They together build a comprehensive recognition context. We construct a series of CPPD models and also plug the proposed modules into existing STR decoders. Experiments on both English and Chinese benchmarks demonstrate that the CPPD models achieve highly competitive accuracy while running approximately 8x faster than their AR-based counterparts. Moreover, the plugged models achieve significant accuracy improvements. Code is at 
\href{https://github.com/PaddlePaddle/PaddleOCR/blob/dygraph/doc/doc_en/algorithm_rec_cppd_en.md}{this https URL}.

\end{abstract}

}
\maketitle
\IEEEdisplaynontitleabstractindextext

\section{Introduction}
\IEEEPARstart{S}{cene} text recognition (STR) aims at reading a character sequence from the text instance cropped from real-world scenes, e.g., text on billboards, packaging boxes, electronic screens, etc. It has attracted attention from a wide range of applications for its key role in precise image interpretation. However, the task still remains challenging due to text variations such as distortion, occlusion, blurring, multi-fonts, etc. Meanwhile, the strict time consumption limit in many applications is another challenge posed to STR models.

\begin{figure}[t] 
\centering
\includegraphics[width=0.35\textwidth]{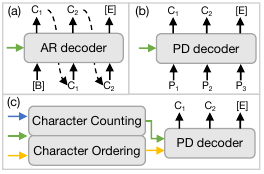}  
\caption{(a) Autoregressive (AR) decoder. (b) Parallel decoding (PD) decoder. (c) Our CPPD. CPPD has two well-designed modules for robust and fast context perception. C and P are character and position embeddings. Green, blue, and yellow arrows denote visual features, character counting, and ordering embeddings, respectively.}  
\label{fig:fig1}  
\end{figure}

Among STR solutions, attention-based encoder-decoder ones are perhaps the most popular paradigm. Typically, these methods first employ a CNN \cite{He_2016_CVPR} or ViT \cite{dosovitskiy2020vit} as the vision encoder. During decoding, they retain all the extracted visual features and dynamically select them for recognition, depending on the decoder used. There are two major decoders, i.e., autoregressive (AR) and parallel decoding (PD) decoders. AR decoders are widely adopted and among the most accurate decoders in NLP tasks like machine translation \cite{NIPS2017_attn, seqtoseq, SMT}, owing to their superior language modeling capability. By using AR decoder for STR, as shown in Fig.~\ref{fig:fig1}(a), the characters are recognized one-by-one and when decoding the \emph{i}-th character, the previously identified \emph{i-1} characters are regarded as priors and fed into the decoder. By doing so, linguistic knowledge is implicitly encoded. However, AR decoders run slowly due to the use of an iterative decoding scheme. Meanwhile, the accuracy reported by earlier AR-based models \cite{li2019sar,Sheng2019nrtr} is worse than recent top-ranked methods~\cite{fang2021abinet,BautistaA22PARSeq}, which is inconsistent with the observations in NLP tasks. Recently, several efforts \cite{wan2020textscanner,yue2020robustscanner,zheng2023cdistnet} have shown that the accuracy of AR decoders could be improved and ranked top-tier by carefully modeling the character position clue. They suggest that language modeling is not the only clue captured by AR decoding.

In contrast, PD decoders infer all the characters in a single forward propagation pass, as shown in Fig.~\ref{fig:fig1}(b). They have received much research attention \cite{yu2020srn,fang2021abinet,Wang_2021_visionlan} due to the advantage in inference speed compared to AR decoders. However, they also encounter the problem that the recognition context could not be stably established in one forward computation. To compensate for this, recent efforts \cite{fang2021abinet,Wang_2021_visionlan} have focused more on better exploiting non-visual clues and have attained impressive recognition accuracy. For example, ABINet \cite{fang2021abinet} introduced an external language model and an iterative correction scheme to aid the recognition. It got highly competitive accuracy in standard benchmarks with a certain sacrifice on inference speed. 

Aiming at inheriting the merits of both AR and PD decoders, i.e., accurate and fast, we first investigate why early AR-based STR models perform inferior. Our exploration reveals that the reason lies in the outdated visual encoder and training strategy. Specifically, NRTR \cite{Sheng2019nrtr}, a Transformer-based AR model, employs a weak encoder. We upgrade it by using SVTR (e.g., SVTR-T) \cite{duijcai2022svtr} and adopting the same training strategy as in \cite{fang2021abinet}. Surprisingly, such a straightforward modification significantly improves the accuracy and makes the generated SVTR-NRTR series top-ranked recognizers in terms of accuracy, which motivates us to choose it as the baseline for further study. We observe that context modeling is more complicated for AR-based STR due to the difference between visual and text modalities. For NLP tasks, what is modeled, in general, is a linguistic context that records the appearance probability of a character given a preceding character sequence. However, for STR, in addition to the linguistic association between characters, visual content such as character appearance and positions can also be different and affect recognition. However, how AR decoders utilize these context variables is still less explored, hindering the design of more advanced recognizers.

To explore this further, we conduct empirical experiments based on SVTR-T-NRTR (i.e., SVTR-T as the encoder). The results suggest that AR decoding outperforms current PD by a relatively large accuracy margin. It is mainly because, in addition to linguistic modeling, AR decoder can get visual context such as the appearance and positions of previously decoded characters, which are absent in current PD. Moreover, when adding a groundtruth-based side loss to PD, it gets accuracy gains but is still worse than AR, implying that a character prediction-guided loss is insufficient to fully infer the context above. Our attempts basically indicate that properly modeling visual context is the key to enhancing the accuracy of PD decoders.

With the exploration above, we focus on developing accurate and efficient STR models from the angle of fast and robustly estimating visual context. To this end, we propose a new PD decoder termed Character Perception Parallel Decoder (CPPD), which aims at simultaneously attaining high accuracy and fast inference speed. As shown in Fig.~\ref{fig:fig1}(c), we devise a character ordering (CO) module and a character counting (CC) module. Both employ cross-attention for targeted feature enhancement. By using dedicated side losses as guidance, CO successfully estimates the content-free placeholders of the characters and their reading order, while CC predicts the occurrence count of each character. Moreover, the character prediction task also guides the learning process to fill in the placeholders with characters. Thus, these components together form a context that besides linguistic modeling, also describes the character count, position, and reading order of the entire character sequence, robustly giving what the character sequence is and where the characters appear. It is worth noting that the context is more complete than that modeled by AR decoding due to the inclusion of all the characters. In addition, since the context above can be established in a single forward propagation, CPPD also accelerates the inference significantly.

Based on the study above, we construct a series of SVTR-CPPD models and also plug the proposed CO and CC modules into existing STR decoders. Extensive experiments on both English and Chinese benchmarks demonstrate that the constructed CPPD models achieve highly competitive accuracy, which is even higher than their AR counterparts while running approximately 8x faster, positioning them among the most accurate and fast recognizers to date. Moreover, the plugged models also improve the accuracy by clear margins, especially on recent large and challenging datasets. The contribution of this paper is threefold. 

\begin{itemize}
    \item We verify that AR decoder remains the most competitive STR decoder in terms of accuracy, and empirically reveal that besides linguistic context, character count, positions, and their reading order are essential context variables. These variables are largely missed in existing PD models.
    \item We devise CPPD, a novel PD decoder that incorporates CO and CC modules with dedicated losses to perceive the desired context in a single forward pass. The two modules and the character recognition task are complementary, and they form a visual and linguistic involved context description.  
    \item Extensive experiments confirm the superiority of CPPD. The constructed models achieve highly competitive accuracy, and run as fast as 400 FPS during inference on one NVIDIA 1080Ti GPU. In addition, integrating CPPD modules into existing STR models leads to significant accuracy improvements.
\end{itemize}

\section{Related Work}
The advance of deep learning has propelled STR into the attention-based encoder-decoder era \cite{chen2021text, YANG20221458}. Generally, the encoder extracts visual, linguistic, and/or other features. The decoder integrates these features and generates the recognized sequence. Based on how the decoder is formulated, we can broadly classify existing models as CTC and attention-based ones, and the latter can be further classified as two categories, i.e., AR and PD models.

CTC models implement the decoding via a two-step alignment, i.e., character-image sub-region mapping and CTC decoding \cite{CTC}. For example, CRNN \cite{shi2017crnn} first extracted the image feature by jointly utilizing CNN and RNN. It then inferred a character or a blank for each image sub-region. The recognized sequence was obtained by de-duplicating characters and removing blanks according to the learned CTC rule. In \cite{hu2020gtc}, GTC employed a graph neural network and attention mechanism to improve both feature learning and CTC decoding. SVTR \cite{duijcai2022svtr} devised a dedicated ViT to capture multi-grained character features and attained impressive accuracy using only CTC decoding. CTC models have fast inference speed in general. Nevertheless, they assume that the recognition is based on character-image sub-region correspondences, restricting their ability to accommodate variations like text curvation, etc.

Inspired by the success of attention-based models in NLP, AR models were introduced to STR in \cite{shi2019aster,li2019sar,wang2020decoupled}, where gated RNN \cite{lstm, gru} was employed for decoding and showed accuracy advantages. NRTR \cite{Sheng2019nrtr} built the first Transformer-based AR decoder \cite{NIPS2017_attn} for STR. It achieved the top accuracy at that time. Later, its accuracy was further improved by using ResNet as the encoder \cite{mmocr2021}. SATRN \cite{lee2020recognizing} was developed based on a 2D Transformer to better preserve 2D features. To suppress attention drift during decoding, character position embedding was leveraged and different position enhancement schemes were proposed in \cite{wan2020textscanner,yue2020robustscanner,zheng2023cdistnet}. They got improved accuracy, implying that the character position was also a critical clue for STR. PARSeq \cite{BautistaA22PARSeq} learned a permuted AR sequence model by exhaustively utilizing the character occurrence dependence. AR decoder was also applied to recognize artistic text in \cite{xie2022toward}. Despite the progress, AR models perform recognition in a character-by-character manner. The inference speed is an issue and hinders their applications in fast-response scenarios. 

PD models were proposed to seek accurate and fast STR. Generally, the speed advantage came from PD while the accuracy was mainly attributed to exploiting non-visual clues so far. For example, SRN \cite{yu2020srn} inferred all the characters at once, where a groundtruth-based side loss was imposed to guide the context modeling. Later, Zhang et al. \cite{zhang2023linguistic} got improved accuracy by recurrently applying such a side loss. In \cite{fang2021abinet}, Fang et al. introduced external language model and the iterative mechanism to its decoder. Wang et al. \cite{Wang_2021_visionlan} devised an additional LM network to endue the vision backbone with LM capability. During inference, the LM network was discarded to ensure speedup. Qiao et al. \cite{qiao2021pimnet} proposed PIMNet. It used a more accurate AR decoder to distill PD decoder. The AR decoder was discarded at inference, achieving a balance between accuracy and efficiency.

Our CPPD also falls into the PD decoder family. It devises two dedicated side loss-guided CC and CO modules to capture the required context. Note that CC-like clues have been exploited in several previous studies. For example, they were explored in the forms of predicting sequence length \cite{jiang2021reciprocal}, or estimating character count using CAN \cite{li2022counting} and ACE \cite{xie2019aggregation}. However, CAN was dedicated to AR-based mathematical expression recognition, and ACE used a probability-based sequence-level loss. Our usage differs from those practices and its detail will be elaborated later. Meanwhile, some works have also explored the position information \cite{yue2020robustscanner,wang2020decoupled,zheng2023cdistnet,yu2020srn,fang2021abinet,Wang_2021_visionlan}. Typically, these references initialized a fixed position embedding and then fused it with other clues to make predictions. These usages are quite different from our side loss-guided CO learning.

\begin{figure}[t] 
\centering
\includegraphics[width=0.47\textwidth]{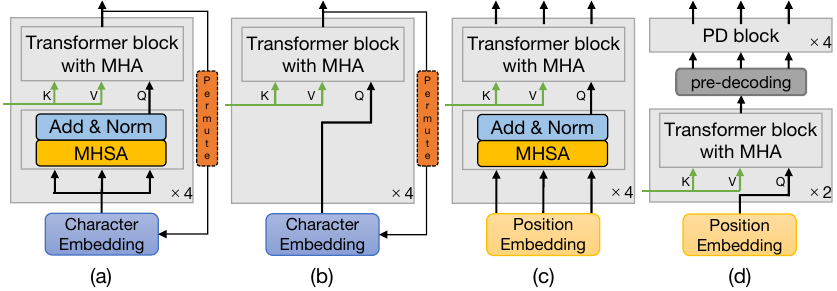}  
\caption{Six decoder variants considered in our empirical studies. (a) AR decoder trained using the left-to-right sequence (AR) and permuted sequence (AR-P, with the dashed red box enabled), (b) AR decoder with limited linguistic context (AR-L) and AR-L trained using permuted sequence (AR-L-P), (c) PD decoder, (d) PD decoder with pre-decoding (PD-P), where PD block is the gray region in (c). MHA (MHSA) denotes multi-head (self-) attention.}  
\label{fig:exp_4}  
\end{figure}

\section{AR Decoder Revisit}
As stated, NRTR \cite{Sheng2019nrtr}, a typical Transformer-based AR model for STR, performed worse than the top-ranked models by some margins. To investigate the reason behind this, we first adopt the same training strategy as in \cite{fang2021abinet} to train NRTR. The obtained model, NRTR-re, gives an average accuracy of 89.62\% on six standard English benchmarks \cite{whatwrong}, surpassing the original NRTR \cite{Sheng2019nrtr} by 2.24\%. We then employ SVTR-T as the visual encoder and retrain NRTR. As seen in Tab.~\ref{tab:sota}, the resulting model, SVTR-T-NRTR, improves the averaged accuracy to 91.49\%, while reducing the model size from 31.7 M to 8.6 M, and increasing the inference FPS from 33.2 to 45.5. Meanwhile, it attains an accuracy improvement of 1.5\% compared to SVTR-T, and an even larger accuracy improvement is observed when SVTR-B is used as the encoder (see Tab.~\ref{tab:sota}), The results highlight the benefits of AR decoding. Note that SVTR-B-NRTR is among the most accurate recognizers, while SVTR-T-NRTR is also the most accurate among models with similar sizes. Since NRTR uses only a standard AR architecture, the results clearly verify the superiority of AR decoding in STR, also indicating that previously inferior accuracy was mainly due to an outdated visual encoder and training strategy.

Then, we investigate the context AR and PD decoders modeled. Specifically, six decoder variants are considered on top of SVTR-T-NRTR. We group them into four categories as shown in Fig.~\ref{fig:exp_4}: (a) standard AR decoder, which is the same as SVTR-T-NRTR, either trained using the standard left-to-right sequence or permuted sequence \cite{BautistaA22PARSeq} (AR-P, with the dashed red box enabled in Fig.~\ref{fig:exp_4}(a)); (b) AR decoder with limited linguistic context (AR-L), where MHSA is removed from the Transformer block, limiting the linguistic modeling only from the AR decoding loop. The same as above, the permuted version is also trained (AR-L-P); (c) Standard PD decoder; (d) PD decoder with pre-decoding (PD-P), where a groundtruth-based side loss the same as \cite{yu2020srn} is added to guide the feature learning. The first four AR-based variants can be uniformly written as:

\begin{align}
    &P(Y| F_v, \theta_D) =  \prod\limits_{t=1}^{T+1} p(y_t|C_t, F_v;  \theta_D)
\end{align}

\noindent where $y_t$ is the character to be inferred at time step $t$. $T$ is the total decoding time steps determined during the decoding process. $y_0$ and $y_{T+1}$ represent the start and end tokens. $C_t$, $F_v$, and $\theta_D$ denote the corresponding recognition context, visual features, and decoder parameters, respectively. In this experiment, these variants only differ from $C_t$, or saying, how the context is formulated. Specifically, they are: 

\begin{align}
    &C_t^{AR} = c_{y_0}:c_{y_{t-1}} \\
    &C_t^{AR-P} = c_{y_{k_0}}:c_{y_{k_{t-1}}} \\
    &C_t^{AR-L} = c_{y_{t-1}} \\
    &C_t^{AR-L-P} = c_{y_{k_{t-1}}}
\end{align}

\noindent where $c_{y_i}$ is the context of character $y_i$, including its category, appearance, position, and linguistic association with other characters. $c_{y_0}:c_{y_{t-1}}$ is the context of previously decoded $t-1$ characters, and $\{k_1, ..., k_{T}\} = Permute\{1,...,T\}$ denotes a permuted sequence.

\begin{table}[t]\footnotesize
\caption{Results of the six decoder
variants.}
\centering
\setlength{\tabcolsep}{3pt}{
\begin{tabular}{c|cccccc}
\hline
Method   & AR    & AR-P & AR-L  & AR-L-P & PD   & PD-P  \\
\hline
Avg(\%)  & 91.49 & 91.78      & 91.13 & 91.45        & 89.60 & 90.43 \\
Time(ms) & 22.01    & 22.01         & 21.19  & 21.19         & 2.630 & 2.801  \\
\hline
\end{tabular}
}

\label{tab:AR}
\end{table}

The average accuracy and time consumption of the six variants are listed in Tab.~\ref{tab:AR}, from which the following empirical observations are inferred. We first look into those from the AR-based ones. First, only 0.36\% and 0.33\% accuracy drops are observed when comparing AR-L with AR, and AR-L-P with AR-P. The result implies that removing the direct language modeling within each time step, which truncates the multigram language model to a bigram one (see Eq. 2 and Eq. 4), does not trigger an obvious accuracy drop. Second, introducing training permutation increases the accuracy by 0.29\% (AR v.s. AR-P) and 0.33\% (AR-L v.s. AR-L-P), as the context has been enriched during the permutation. Note that AR-L-P attains similar accuracy as the standard AR although it depends on only one permuted character context, i.e., with a weaker linguistic context. The result indicates that other context variables also have been strengthened during the training permutation.

\begin{figure}[t] 
\centering
\includegraphics[width=0.47\textwidth]{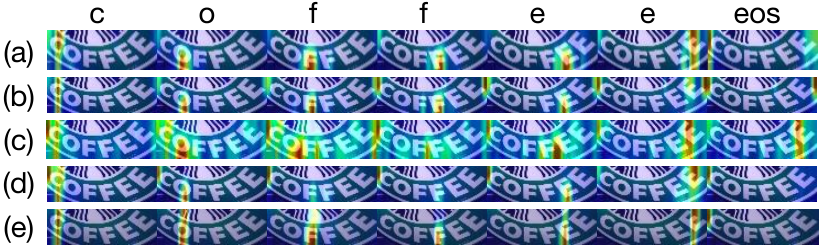}  
\caption{Attention maps of different decoder variants. (a) AR decoder, (b) AR-L, (c) PD decoder, (d) PD-P, (e) our CPPD.}  
\label{fig:vis1}  
\end{figure}

We then compare AR with the PD ones. First, when changing from the AR decoder to a standard PD decoder, despite accelerating inference speed, a 1.89\% accuracy drop is observed. As seen in Fig.~\ref{fig:exp_4}(c), the PD decoder uses a fixed content-free position embedding as the \emph{query} to associate with visual features, which may be quite different across text instances. Thus, it is challenging to establish a robust context in such a forward pass. This is also reflected in the attention maps in Fig.~\ref{fig:vis1}(c), where the \emph{query} could not accurately gaze at the corresponding characters. Second, when pre-decoding is added, PD-P gains an accuracy improvement of 0.83\% compared to the standard one. Basically, this loss imposes an additional learning objective that also uses groundtruth characters as the aligning target. It drives the STR model to separate context modeling and character prediction tasks to some extent, thus enjoying more accurate character localization, as shown in Fig.~\ref{fig:vis1}(d). However, it is observed that context variables like character appearance and positions still have not been explicitly modeled. This explains why there remains a margin between AR and PD-P decoders.

The experiments basically verify that compared to PD decoder, the success of AR decoder mainly lies in the availability of the previously decoded characters, whose appearance and positions serve as necessary visual context and aid the recognition. However, AR decoder also causes a slow inference speed. How to appropriately model those context variables in a PD decoding pass, and achieve the goal of "AR's accuracy, PD's speed" still needs exploration.

\begin{figure*}[ht]  
\centering  
\includegraphics[width=0.85\textwidth]{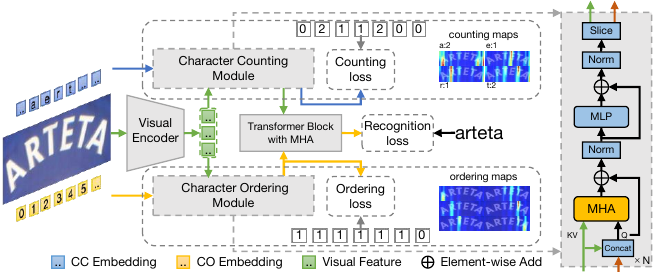}  
\caption{An overview of CPPD. The decoder accepts visual features, character counting and ordering embeddings. Cross-attention and corresponding side losses are equipped to guide model learning. The right side shows details of the character counting and ordering modules, where the red line denotes either character counting or ordering embeddings, depending on which module is employed.} 
\label{fig:structrue}  
\end{figure*}

\section{The Proposed CPPD}
\label{sec:formatting}

\subsection{Overall}
We propose CPPD to address the above issue. Fig.~\ref{fig:structrue} gives an overview of CPPD. Given a text instance of $I \in R^{H \times W \times 3}$, we first use SVTR \cite{duijcai2022svtr} (with the rectification module and CTC decoder removed) as the encoder to extract visual feature $\mathbf{F}_v \in R^{ \frac{H}{16} \times \frac{W}{4} \times D}$. Meanwhile, character counting (CC) embedding $\mathbf{E}_{cc} \in R^{C \times D}$ and character ordering (CO) embedding $\mathbf{E}_{co} \in R^{L \times D}$ are initialized with a truncated normal distribution with mean of 0 and standard deviation of 0.2, where $C$ and $L$ are predefined parameters denoting the size of character set and the maximum length of a character sequence, respectively. On the decoder side, a CC module and a CO module are developed, which use either $\mathbf{E}_{cc}$ or $\mathbf{E}_{co}$ as \emph{query}, $\mathbf{F}_v$ as \emph{key} and \emph{value}, to calculate cross-attention. Two dedicated side losses are devised to guide their learning. Their output features are further fused by another cross-attention. As a result, it generates the desired context features. Then a character recognition task is followed, which uses cross-entropy (CE) loss to learn the decoded sequence in a single forward propagation. We detail the structure of CPPD as follows. 

\subsection{Character counting and ordering modules}
CC and CO modules are devised for feature reinforcement and context modeling. The two modules have the same internal structure but their optimization objectives are different. The CC module is followed by a character counting (CC) loss, which aims at describing the occurrence count of each character as accurately as possible. For example, given \emph{arteta}, the task is to tell us that character \emph{a} and \emph{t} both appear twice, character \emph{e} and \emph{r} appear once, while other characters do not appear, i.e., with count 0. By forcing this, the CC embedding $\mathbf{E}_{cc}$, which is used as \emph{query}, can search for the desired information in visual feature space. Since different characters have different appearances and two characters with the same category are likely to have similar feature representations, the \emph{query} feature can be strengthened from this dimension. 

Let $y_{c,l}=1$ if the \emph{c}-th character appears \emph{l} times in the character sequence and 0 otherwise, $p_{c,l}$ the probability that the \emph{c}-th character be predicted to appear \emph{l} times in the same sequence. The CC loss $\mathcal{L}_{cc}$ is written as:
\begin{equation}
\mathcal{L}_{cc} = -\frac{1}{C}\sum_{c=1}^C\sum_{l=0}^Ly_{c,l}\log(p_{c,l})
\label{equ:ce}
\end{equation}

Note that Eq.~\ref{equ:ce} assigns a loss term for every possible character and occurrence frequency. Compared with the ACE loss (Eq.~\ref{equ:ace}) \cite{xie2019aggregation} that only considered the appeared characters, our CC loss has the merits of more completely modeling the occurrence of characters and being decoupled from the character sequence length \emph{L}. We also have conducted an experiment comparing the two (see Tab.~\ref{tab:loss}).

\begin{equation}
\mathcal{L}_{ace} = -\sum_{c=1}^C \frac{N_{c}}{L}\log(\frac{\sum_{l=0}^Lp_{c,l}}{L})
\label{equ:ace}
\end{equation}

Meanwhile, the CO module is followed by a character ordering (CO) loss, which aims at inferring the locations from the first character to the last one. Also taking \emph{arteta} as an example, it focuses on deducing where the first to sixth characters appear regardless of their categories, i.e., the content-free reading order and positions. Similar to above, the CO embedding $\mathbf{E}_{co}$ is treated as \emph{query} to implement this goal. Since the character foreground exhibits differences with the background, features discriminating character positions could be identified and extracted from the visual feature space, and used to reinforce the \emph{query} feature.

Let $p_{l}$ be the probability that the \emph{l}-th position is predicted to have a character. The CO loss $\mathcal{L}_{co}$ is given by:
\begin{equation}
\mathcal{L}_{co} =  -\frac{1}{L}\sum_{l=0}^L{(y_{l}\log(p_{l}) + (1 - y_{l})\log(1 - p_{l}))}
\label{equ:bce}
\end{equation}

\noindent where $y_{l}=1$ for $l\leq N$ and $y_{l}=0$ for $l> N$ always holds for a sequence of length \emph{N}. $\mathcal{L}_{co}$ is binary CE loss over all the \emph{L} positions. Note that the same label will be assigned to two different character sequences with the same length, therefore forcing CO to capture the content-free placeholders. Meanwhile, since different sequences may have different lengths and they are all initialized in a left-to-right order. This loss also implicitly encodes the reading order. Consequently, it decouples the CO task from the final character prediction task to some extent, forming a definite CO enhancement. It is worth mentioning that such a usage is quite different from existing PD models \cite{yu2020srn,fang2021abinet,Wang_2021_visionlan}.

We take the CC module as an example to illustrate the detailed implementation of the two modules. As depicted on the right side of Fig.~\ref{fig:structrue}, the visual feature $\mathbf{F}_v$ is first reshaped and concatenates with the CC feature $\mathbf{E}_{cc}$, forming the \emph{query} feature and experienced an MHA. Then layer norm and MLP are successively carried out. Residual connections are also employed for feature consolidation. Next, the feature is sliced into a new visual feature and a new CC feature with the same shape as the inputted ones. The process is performed twice (for SVTR-T) or three times (for SVTR-B), depending on the visual encoder employed. The generated features are termed $\mathbf{F}_{cc}$ and $\mathbf{\hat{E}}_{cc}$ for visual and CC, respectively. Similarly, $\mathbf{F}_{co}$ and $\mathbf{\hat{E}}_{co}$ can be obtained the same as above from the CO module.

\subsection{Recognition module}
The two modules above generate strengthened features biased towards the occurrence count of each character and the content-free positions. They are different but complementary. Therefore, we fuse them using another cross-attention.

Specifically, $\mathbf{\hat{E}}_{co}$ is recruited as \emph{query}, and $\mathbf{F}_{cc}$ is treated as \emph{value} and \emph{key}. A standard Transformer block with MHA is employed to conduct the cross-attention. It generates a combined feature that better characterizes the context. Finally, we implement the recognition by using a simple parallel linear prediction, where cross-entropy (CE)-based recognition loss as in Eq.~\ref{equ:ce2} is imposed to learn the correct decoding. The reason that we uses $<\mathbf{\hat{E}}_{co},\mathbf{F}_{cc}>$ pair rather than $<\mathbf{\hat{E}}_{cc},\mathbf{F}_{co}>$ is twofold. First, $\mathbf{\hat{E}}_{co}$ encodes the reading positions, which is more in line with the character prediction task. Second, the size of the character set $C$ is large for languages like Chinese, using $\mathbf{\hat{E}}_{cc}$ would consume more computational resources. 

\begin{equation}
\mathcal{L}_{rec} =  -\frac{1}{L}\sum_{l=0}^{L}\sum_{c=1}^Cy_{l,c}\log(p_{l,c})
\label{equ:ce2}
\end{equation}

\noindent where $p_{l,c}$ is the probability that the \emph{l}-th character in the sequence being classified as character \emph{c}. $y_{l,c}=1$ if character \emph{c} is the label of the \emph{l}-th character, and 0 otherwise.

\subsection{Training Objective}

CPPD is trained in an end-to-end manner with the CC loss in Eq.~\ref{equ:ce}, CO loss in Eq.~\ref{equ:bce} and CE loss in Eq.~\ref{equ:ce2}. The final optimization function is formulated as:

\begin{equation}
    \mathcal{L} = \lambda_{co}\mathcal{L}_{co} + \lambda_{cc}\mathcal{L}_{cc}+ \lambda_{rec}\mathcal{L}_{rec}
\label{eq:all}
\end{equation}

\noindent where $\lambda_{co}$, $\lambda_{cc}$ and $\lambda_{rec}$ are all simply set to 1 in ours experiments. 

The CE loss focuses on a position-irrelevant character prediction. Together with the CC and CO losses, they form a mutually reinforced recognition context, which associates the content-free placeholders with position-free characters. Thus, the constructed context robustly delineates character content, count, positions, and their reading order, forming a comprehensive description of character appearance and positions. Meanwhile, Linguistic context is also learned in the form of character appearing dependency from the CC and the final recognition modeling. As a result, CPPD is capable of building a visual and linguistic involved context, which is even more complete than that of AR decoding due to the inclusion of subsequent characters. Benefiting from their joint optimization, as shown in Fig.~\ref{fig:structrue}, visualization on the CC side predicts the character counts and their positions accurately. While on the CO side, it clearly captures the character's reading order and positions. Both of these explain the merit of CPPD from the loss perspective. 

\section{Experiments}

\subsection{Datasets and Implementation Details}
We validate CPPD on both English and Chinese datasets. For English our models are trained on two commonly used synthetic scene text datasets, i.e., MJSynth (MJ) \cite{jaderberg14synthetic, Jader2015Reading} and SynthText (ST) \cite{Synthetic}. Then the models are tested on: (1) six regular and irregular text benchmarks, i.e., ICDAR 2013 (IC13) \cite{icdar2013}, Street View Text (SVT) \cite{Wang2011SVT}, IIIT5K-Words (IIIT) \cite{IIIT5K}, ICDAR 2015 (IC15) \cite{icdar2015}, Street View Text-Perspective (SVTP) \cite{SVTP} and CUTE80 (CUTE) \cite{Risnumawan2014cute}. For IC13 and IC15, we use the versions with 857 and 1,811 images, respectively. (2) the recently constructed Union14M-L benchmark, which consists of over 0.4 million real-world test images with both complexity and versatility \cite{jiang2023revisiting}. It includes challenging text subsets such as curve, multi-oriented, artistic, etc. 

For Chinese we use Chinese text recognition (CTR) benchmark \cite{chen2021benchmarking}, a public dataset containing four challenging scenarios: Scene, Web, Document and Writing. For each scenario, we train the model on the corresponding training set and use the validation set to determine it, which is then assessed on the test set.

We use AdamW optimizer \cite{adamw} with a weight decay of 0.05 for training. For English models, all text instances, unless specified, are resized to $32 \times 100$ and the learning rate (LR) is set to $\frac{5}{10^{4}} \times \frac{batchsize}{1024}$. Cosine LR scheduler \cite{cosine} with 4 epochs linear warm-up is used in all the 20 epochs. 
Data augmentation like rotation, perspective distortion, motion blur and gaussian noise, are randomly performed during training. The alphabet includes all case-insensitive alphanumerics. For Chinese models, all text instances are resized to $32 \times 256$. Data augmentation is not performed and the LR is set to $\frac{5}{10^{4}} \times \frac{batchsize}{512}$. Cosine LR scheduler with 5 epochs linear warm-up is used in all the 100 epochs. Word accuracy is used as the evaluation metric. The size of the character set $C$ is set to 37 for English and 6625 for Chinese \cite{ppocrv3}. The maximum prediction length $L$ is set to 25 for both. All models are trained on 4 Tesla V100 GPUs.

\subsection{Ablation Study} \label{section:3.3}
To better understand CPPD, we perform controlled experiments under different settings on SVTR-B-CPPD.

\begin{table}[t]\footnotesize
\caption{Ablation study on loss choice of CC and CO.}
\centering
\setlength{\tabcolsep}{4pt}{
\begin{tabular}{c|c|c|c}
\hline

\multicolumn{2}{c|}{Loss}  & Counting or ordering  label of \textit{arteta}        & Avg(\%) \\
\hline
\multirow{2}{*}{Ours}        & Eq.~\ref{equ:ce}        & {[}a:2,e:1,r:1,t:2,other:0{]} & \multirow{2}{*}{93.80}        \\
 & Eq.~\ref{equ:bce}        & {[}1,1,1,1,1,1,0,…,0{]}   &         \\
 \hline
ACE        & Eq.~\ref{equ:ace}        & {[}a:$\frac{2}{L}$,e:$\frac{1}{L}$,r:$\frac{1}{L}$,t:$\frac{2}{L}$, \textit{pad}:$\frac{L-6}{L}${]} & 93.31       \\

CE        & Eq.~\ref{equ:ce2}        & {[}a,r,t,e,t,a,[E],\textit{pad},...,\textit{pad}{]} & 93.50       \\
\hline
\end{tabular}}

\label{tab:loss}
\end{table}

\begin{table}[t]\footnotesize
\centering
\caption{Ablation study on CC and CO modules.}
\setlength{\tabcolsep}{2pt}{
\begin{tabular}{cccc|cccccc|c}
\hline
CO & \begin{tabular}[c]{@{}c@{}}CO\\ Loss\end{tabular} & CC & \begin{tabular}[c]{@{}c@{}}CC\\ Loss\end{tabular} & IC13          & SVT           & IIIT & IC15          & SVTP          & CUTE & Avg(\%)   \\
\hline
          &                                                   &           &                                                   & 97.3          & 94.4          & 96.2 & 86.4          & 89.1          & 88.5 & 92.01 \\
\checkmark &                                                   &           &                                                   & 97.3          & 95.2          & 97.2 & 86.9          & 90.2          & 91.0 & 92.97 \\
\checkmark & \checkmark                                         &           &                                                   & 97.5          & 94.4          & 96.9 & 87.2          & 90.5          & \textbf{93.1} & 93.29 \\
          &                                                   & \checkmark &                                                   & 97.2          & 95.1          & 97.2 & 87.0          & 89.3          & 89.2 & 92.49 \\
          &                                                   & \checkmark & \checkmark                                         & 97.4          & 95.2          & 97.1 & 87.3          & \textbf{91.0}          & 92.0 & 93.35 \\
\checkmark &                                                   & \checkmark &                                                   & 97.3          & 95.4          & 97.0 & 87.3          & 90.4          & 91.0 & 93.05 \\
\checkmark & \checkmark                                         & \checkmark &                                                   & 97.3          & 94.6          & 97.1 & 87.0          & 90.7          & 92.4 & 93.18 \\
\checkmark &                                                   & \checkmark & \checkmark                                         & 98.0          & 94.9          & 97.4 & 87.7          & 90.7          & 92.4 & 93.52 \\
\checkmark & \checkmark                                         & \checkmark & \checkmark                                         & \textbf{98.2} & \textbf{95.5} & \textbf{97.6} & \textbf{87.9} & 90.9 & 92.7 & \textbf{93.80}
\\

\hline
\end{tabular}}

\label{tab:co}
\end{table}

\begin{table}[t]\footnotesize
\caption{Feature representativeness analysis of different CPPD components.}
\centering
\setlength{\tabcolsep}{2pt}{
\begin{tabular}{l|cccccc|c}
\hline
Features        & IC13  & SVT   & IIIT  & IC15  & SVTP  & CUTE  & Avg(\%)   \\
\hline
$\mathbf{F}_{cc}$ & 96.3 & 93.7 & 95.9 & 85.9 & 88.5 & 90.3 & 91.76 \\
$\mathbf{F}_{co}$ & 95.7 & 92.0 & 95.0 & 85.1 & 85.3 & 86.8 & 90.00 \\
$\mathbf{F}_{v}$  & 95.2 & 88.4 & 93.4 & 81.5 & 83.3 & 86.8 & 88.10\\

$\mathbf{\hat{F}}_{v}$  & 94.7 & 91.5 & 94.5 & 83.3 & 84.7 & 90.3 & 89.83\\
\hline
\end{tabular}}

\label{tab:ccfeatures}
\end{table}

\begin{table}[t]\footnotesize
\centering
\caption{Results on adaptability to different encoders.}
\setlength{\tabcolsep}{2pt}{
\begin{tabular}{r|c|cccccc|c}
\hline
Decoder & Encoder                   & IC13  & SVT   & IIIT  & IC15  & SVTP  & CUTE  & Avg(\%)   \\
\hline
CTC    & \multirow{3}{*}{ResNet+En} & 96.7 & 93.4 & 96.6 & 85.0 & 86.8 & 91.3 & 91.64 \\
ABINet \cite{TPAMI2022ABINetPP} &                            & 97.4 & 93.5 & 96.2 & 86.0 & 89.3 & 89.2 & 91.93 \\
CPPD   &                            & 97.0 & 94.3 & 96.8 & 87.3 & \textbf{91.2} & \textbf{94.1} & 93.43 \\
\hline
CTC    & \multirow{3}{*}{ViT}       & 96.4 & 93.8 & 96.2 & 86.5 & 87.9 & 91.0 & 91.97 \\
PARSeq \cite{BautistaA22PARSeq} &                            & 97.0 & 93.6 & 97.0 & 86.5 & 88.9 & 92.2 & 92.53 \\
CPPD   &                            & 97.0 & 94.3 & 97.1 & 88.1 & 90.2 & 92.4 & 93.17 \\
\hline
CTC    & \multirow{5}{*}{SVTR-Base}    & 97.1 & 91.5 & 96.0 & 85.2 & 89.9 & 91.7 & 91.90 \\
NRTR &                            & 97.7 & 95.4 & 97.4 & \textbf{88.5} & 90.7 & 91.7 & 93.54 \\
ABINet &                            & 97.0 & 94.4 & 97.2 & 87.5 & 90.1 & 92.0 & 93.02 \\
PARSeq &                            & 97.3 & 95.2 & 96.7 & 87.1 & 90.2 & 91.0 & 92.93 \\
CPPD   &                            & \textbf{98.2} & \textbf{95.5} & \textbf{97.6} & 87.9 & 90.9 & 92.7 & \textbf{93.80} \\
\hline
\end{tabular}}

\label{tab:encoder3}
\end{table}

\begin{figure*}[t] 
\centering
\includegraphics[width=0.98\textwidth]{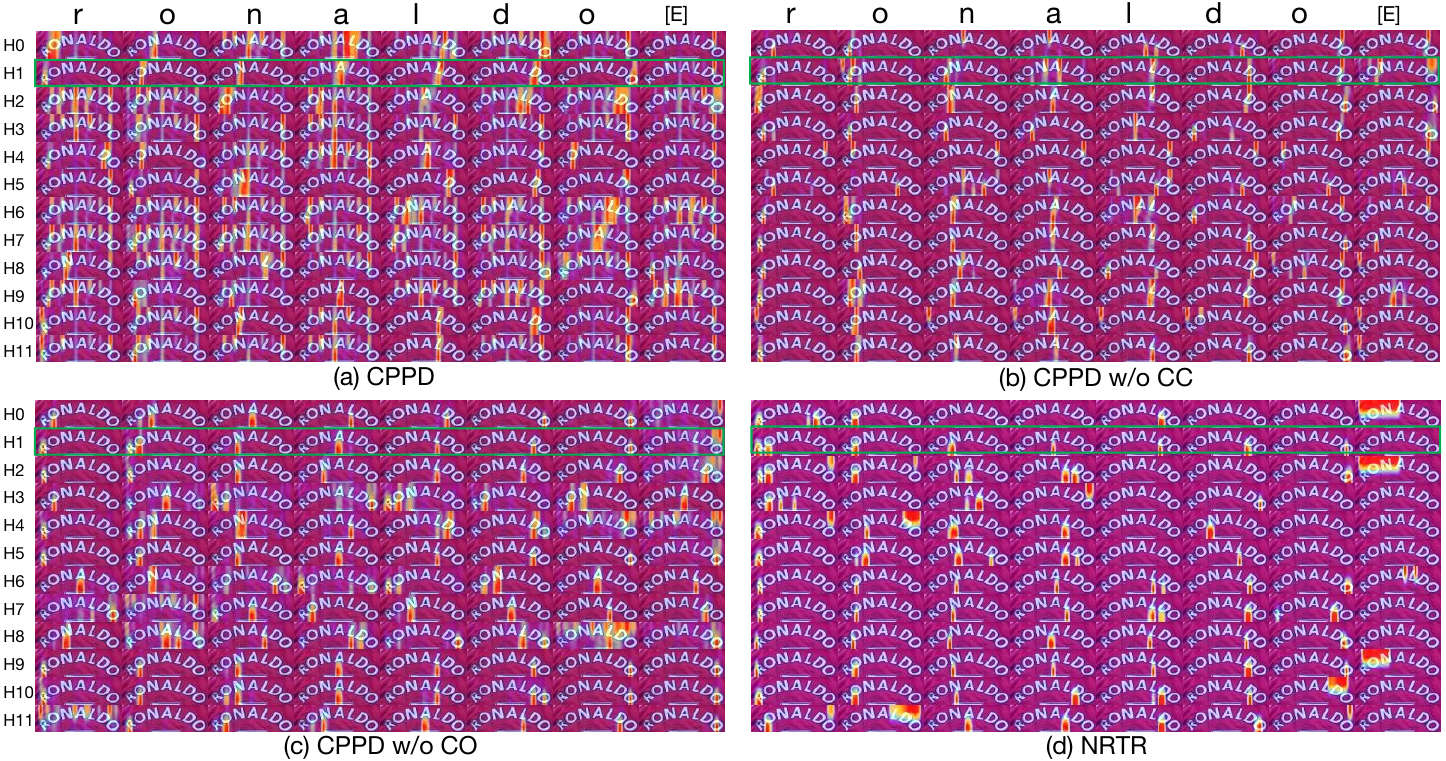}  
\caption{Attention maps of the final recognition features for different models. (a) CPPD, (b) CPPD w/o CC, (c) CPPD w/o CO, (d) NRTR. Each sub-figure depicts the 12 heads in MHSA from top to bottom, and each map corresponds to a character. Best viewed in color.}  
\label{fig:cc_co_vis1}
\end{figure*}

\begin{table*}[t]\footnotesize
\centering
\caption{Results on the six English benchmarks tested against existing methods. NRTR-re denotes the re-trained NRTR by adopting the same training strategy as in \cite{fang2021abinet}. FPS is obtained by averaging the inference time over 3000 English text instances on one NVIDIA 1080Ti GPU under the PaddlePaddle inference mode. The same below.}
\setlength{\tabcolsep}{4pt}{
\begin{tabular}{c|r|c|c|ccc|ccc|c|cc}
\hline
\multicolumn{2}{c|}{\multirow{2}{*}{Method}} & \multicolumn{1}{c|}{\multirow{2}{*}{Encoder}}  & \multicolumn{1}{c|}{\multirow{2}{*}{Input Size}}  & \multicolumn{3}{c|}{Regular}                              & \multicolumn{3}{c|}{Irregular}                            & \multirow{2}{*}{\begin{tabular}[c]{@{}c@{}}Average\\ Accuracy(\%)\end{tabular}} & \multirow{2}{*}{\begin{tabular}[c]{@{}c@{}}Parameters\\ ($\times10^6$)\end{tabular}} & \multirow{2}{*}{\begin{tabular}[c]{@{}c@{}} FPS\end{tabular}} \\
\multicolumn{2}{c|}{}  & \multicolumn{1}{c|}{} & \multicolumn{1}{c|}{}                                                    & IC13          & SVT           & \multicolumn{1}{c|}{IIIT} & IC15          & SVTP          & \multicolumn{1}{c|}{CUTE} &                                                                             &                                                                         &                                                                      \\
\hline
\multirow{6}{*}{AR}     & ASTER \cite{shi2019aster} (2019)    & ResNet    & $32\times100$       &               & 89.5          & 93.4          & 76.1          & 78.5          & 79.5          &                                                                             & 27.2                                                                    &                                                                      \\
                        & NRTR \cite{Sheng2019nrtr} (2019)   & CNN+En    & $32\times100$          & 95.8          & 91.5          & 90.1          &     79.4          &   86.6            &     80.9          &                                                       87.38                      & 31.7                                                                    & 33.2                                                              \\
                        & NRTR-re\cite{Sheng2019nrtr} (2019)& CNN+En & $32\times100$            & 96.6          & 91.8          & 93.0          & 83.4          & 86.8          & 86.1          & 89.62                                                                       & 31.7                                                                    & 33.2                                                              \\
                       
                        & SAR \cite{li2019sar} (2019)& ResNet     & $48\times160$           & 91.0          & 84.5          & 91.5          & 69.2          & 76.4          & 83.5          & 82.68                                                                       & 57.7                                                                    & 9.62                                                             \\
                        & RobustScanner \cite{yue2020robustscanner} (2020)& ResNet      & $48\times160$   & 94.8          & 88.1          & 95.3          & 77.1          & 79.5          & 90.3          & 87.52                                                                       &                 48.0                                                        &  16.4                                                                    \\
                        
                        & PARSeq  \cite{BautistaA22PARSeq} (2022)  & ViT        & $32\times128$        & 97.0          & 93.6          & 97.0 & 86.5          & 88.9          & 92.2          & 92.53                                                                       & 23.8                                                                    & 84.7     \\
                        & CDistNet  \cite{zheng2023cdistnet} (2023)  & ResNet+En        & $32\times128$        & 97.4          & 93.5          & 96.4 & 86.0          & 88.7          & 93.4          & 92.57                                                                       & 65.5                                                                    & 14.5     \\
\hline
\multirow{6}{*}{PD}     & SRN \cite{yu2020srn} (2020)  & ResNet+FPN     & $64\times256$           & 95.5          & 91.5          & 94.8          & 82.7          & 85.1          & 87.8          & 89.57                                                                       & 54.7                                                                    & 39.4                                                                 \\
                        & VisionLAN \cite{Wang_2021_visionlan} (2021)& ResNet+En       & $64\times256$      & 95.7          & 91.7          & 95.8          & 83.7          & 86.0          & 88.5          & 90.23                                                                       & 32.8                                                                    & 200                                                                    \\
                        & ABINet \cite{TPAMI2022ABINetPP} (2021) & ResNet+En     & $32\times128$          & 97.4          & 93.5          & 96.2          & 86.0          & 89.3          & 89.2          & 91.93                                                                       & 36.7                                                                    & 124                                                               \\
                        & GTR \cite{HeC0LHWD22GTR} (2022) & ResNet       & $64\times256$           & 96.8          & 94.1          & 95.8          & 84.6          & 87.9          & 92.3          & 91.92                                                                       & 42.1                                                                    &    \\
                        & I2C2W \cite{tpami2023I2C2W} (2023) & ResNet+En       & $64\times600$           & 95.0          & 91.7          & 94.3          & 82.8          & 83.1          & 93.1          & 90.00                                                                       &                                                                     & \\
                        & LPV-B \cite{ijcai2023LPV} (2023) & SVTR-Base       & $48\times160$           & 97.6          & 94.6          & 97.3          & 87.5          & 90.9          & 94.8          & 93.78                                                                       &   35.1     & 103 \\
\hline
\multirow{4}{*}{CTC}    & CRNN \cite{shi2017crnn} (2017)& ResNet+BiLSTM     & $32\times100$           & 91.1          & 81.6          & 82.9          & 69.4          & 70.0          & 65.5          & 76.75                                                                       & 8.30                                                                     & 159                                                                  \\
                        & SVTR-T \cite{duijcai2022svtr} (2022)& SVTR-Tiny    & $32\times100$           & 96.3          & 91.6          & 94.4          & 84.1          & 85.4          & 88.2          & 89.99                                                                       & 6.03                                                                    & 408                                                               \\
                        & SVTR-B \cite{duijcai2022svtr} (2022) & SVTR-Base     & $48\times160$        & 97.1          & 91.5          & 96.0          & 85.2          & 89.9          & 91.7          & 91.90                                                                       & 24.6                                                                    & 212                                                               \\
                        & SVTR-L \cite{duijcai2022svtr} (2022)& SVTR-Large     & $48\times160$         & 97.2          & 91.7          & 96.3          & 86.6          & 88.4          & \textbf{95.1}          & 92.54                                                                       & 40.8                                                                    & 161                                                                \\
\hline
\multirow{5}{*}{Ours}    & SVTR-T-NRTR & SVTR-Tiny & \multirow{4}{*}{$32\times100$}    & 97.3          & 93.4          & 95.0          & 85.5          & 88.2          & 89.6          & 91.49                                                                       & 8.56                                                                     & 45.5                                                              \\
                        & SVTR-B-NRTR  & SVTR-Base  &      & 97.7 & 95.4 & 97.4 & 88.5 & 90.7 & 91.7 & 93.54                                                              & 32.2                                                                    & 32.5                                                               \\
                
                        & SVTR-T-CPPD & SVTR-Tiny  &     & 97.1          & 94.4          & 96.6          & 86.6 & 88.5          & 90.3          & 92.25                                           & 8.29                                                                      & 399                                                               \\
                        & SVTR-B-CPPD  & SVTR-Base &     & 98.2 & 95.5 & 97.6 & 87.9 & 90.9 & 92.7 & 93.80                                                                       & 26.8                                                                      & 252                                                               \\
\cline{4-4}                        & SVTR-B-CPPD  & SVTR-Base &   $48\times160$  & 97.5 & 95.5 & \textbf{97.7} & 87.7 & 92.4 & 93.7 & 94.10                                                                      & 26.8                                                                      & 206                                                               \\
\hline
\multirow{4}{*}{Plug}    & CTC-CPPD & \multirow{4}{*}{SVTR-Base} & \multirow{4}{*}{$32\times100$}     & 96.8 & 95.2 & 96.9 & 87.2 & 90.2 & 91.7 & 93.02 & 24.6 &212\\
&NRTR-CPPD &     &                        & \textbf{98.6} & \textbf{96.6} & 97.6 & \textbf{88.8} & 91.9 & 93.4 & \textbf{94.50} & 32.3 & 31.1\\
&ABINet-CPPD &       &                      & 98.0 & 96.3 & 97.2 & 87.6 & \textbf{92.6} & 92.4 & 94.01 & 38.9 & 141\\
&PARSeq-CPPD &               &              & 97.3 & 95.7 & 97.4 & 87.6 & 90.9 & 93.4 & 93.71 & 24.2 &98.2\\
                        \hline
\end{tabular}
}

\label{tab:sota}
\end{table*}

\begin{table*}[t]\footnotesize
\centering
\caption{Union14M-L performance of models trained on \textbf{synthetic datasets} (MJ and ST). }
\setlength{\tabcolsep}{3pt}{
\begin{tabular}{c|r|c|c|ccccccc|c}
\hline
\multicolumn{2}{c|}{Method} & Encoder & Input Size  & Curve & \begin{tabular}[c]{@{}c@{}}Multi-\\ Oriented\end{tabular} & Artistic & Contextless & Salient & \begin{tabular}[c]{@{}c@{}}Multi-\\ Words\end{tabular} & General & \begin{tabular}[c]{@{}c@{}}Average\\  Accuracy(\%)\end{tabular}   \\
\hline
\multirow{5}{*}{AR}   & ASTER \cite{shi2019aster} (2019)  & ResNet      & $32\times100$    & 34.0                 & 10.2                                                      & 27.7                 & 33.0                 & 48.2                 & 27.6                                                   & 39.8                 & 31.50                                             \\
                      & NRTR \cite{Sheng2019nrtr} (2019)  & CNN+En      & $32\times100$     & 31.7                 & 4.4                                                       & 36.6                 & 37.3                 & 30.6                 & 54.9                                                   & 48.0                 & 34.79                                             \\
                      & SAR \cite{li2019sar} (2019)  & ResNet   & $48\times160$       & 44.3                 & 7.7                                                       & 42.6                 & 44.2                 & 44.0                 & 51.2                                                   & 50.5                 & 40.64                                             \\
                      & RobustScanner \cite{yue2020robustscanner} (2020)& ResNet  & $48\times160$  & 43.6                 & 7.9                                                       & 41.2                 & 42.6                 & 44.9                 & 46.9                                                   & 39.5                 & 38.09                                             \\
                      
                      & PARSeq-B \cite{BautistaA22PARSeq} (2022) & SVTR-Base   & $32\times100$   & 63.9                 & 16.7                                                      & 52.5                 & 54.3                 & 68.2                 & 55.9                                                   & 56.9                 & 52.62                                             \\
                      & CDisNet \cite{zheng2023cdistnet} (2023) & ResNet+En   & $32\times128$   & 69.3                 & 24.4                                                      & 49.8                 & 55.6                 & 72.8                 & 64.3                                                   & 58.5                 & 56.38                                             \\
                      \hline
\multirow{4}{*}{PD}   & SRN \cite{yu2020srn} (2020)   & ResNet+FPN   & $64\times256$      & 63.4                 & 25.3                                                      & 34.1                 & 28.7                 & 56.5                 & 26.7                                                   & 46.3                 & 40.14                                             \\
                      & VisionLAN \cite{Wang_2021_visionlan} (2021) & ResNet+En   & $64\times256$   & 57.7                 & 14.2                                                      & 47.8                 & 48.0                 & 64.0                 & 47.9                                                   & 52.1                 & 47.39                                             \\
                      & ABINet  \cite{TPAMI2022ABINetPP} (2022) & ResNet+En  & $32\times128$     & 59.5                 & 12.7                                                      & 43.3                 & 38.3                 & 62.0                 & 50.8                                                   & 55.6                 & 46.03                                             \\
                      & ABINet-B \cite{TPAMI2022ABINetPP} (2022) & SVTR-Base  & $32\times100$     & 62.3                 & 13.9                                                      & 50.0                 & 45.1                 & 67.1                 & 53.4                                                   & 58.5                 & 50.07                                             \\
                      & LPV-B \cite{ijcai2023LPV} (2023) & SVTR-Base  & $48\times160$     & 68.3                 & 21.0                                                      & 59.6                 & 65.1                 & 76.2                 & 63.6                                                   & 62.0                 & 59.40                                             \\
                      \hline
\multirow{4}{*}{CTC}  & CRNN  \cite{shi2017crnn} (2017)  & ResNet+BiLSTM  & $32\times100$       & 7.5                  & 0.9                                                       & 20.7                 & 25.6                 & 13.9                 & 25.6                                                   & 32.0                 & 18.03                                             \\
                      & SVTR-T \cite{duijcai2022svtr} (2022) & SVTR-Tiny   & $32\times100$      & 63.0                 & 32.1                                                      & 37.9                 & 44.2                 & 67.5                 & 49.1                                                   & 52.8                 & 49.51                                             \\
                      & SVTR-B \cite{duijcai2022svtr} (2022) & SVTR-Base   & $48\times160$      &    69.8                  &   37.7                                                        &  47.9                    &     61.4                 &   66.8                   &   44.8                                                     &  61.0                    & 55.63                                                  \\
                      & SVTR-L \cite{duijcai2022svtr} (2022)  & SVTR-Large   & $48\times160$      &    \textbf{76.8}     &   \textbf{45.2}   &  55.2  &  63.0  &   73.8     &    48.9  & 64.2    &   61.01                                                \\
                      \hline
\multirow{5}{*}{Ours}  & SVTR-T-NRTR   & SVTR-Tiny   & \multirow{4}{*}{$32\times100$}     & 60.8                 & 23.2                                                      & 46.4                 & 43.6                 & 59.0                 & 38.7                                                   & 55.6                 & 46.76                                             \\
                      & SVTR-B-NRTR   & SVTR-Base  &   & 66.2                 & 19.4                                                      & 54.0                 & 58.8                 & 70.5                 & 61.3                                                   & 65.3                 & 56.49                                             \\
                      & SVTR-T-CPPD   & SVTR-Tiny   &  & 52.4                 & 12.3                                                      & 48.2                 & 54.4                 & 61.5                 & 53.4                                                   & 61.4                 & 49.10                                             \\
                      & SVTR-B-CPPD  & SVTR-Base    &  & 65.5                 & 18.6                                                      & 56.0                 & 61.9                 & 71.0                 & 57.5                                                   & 65.8                 & 56.63                                             \\
\cline{4-4} & SVTR-B-CPPD  & SVTR-Base &   $48\times160$  & 71.9 & 22.1 & \textbf{60.5} & \textbf{67.9} & \textbf{78.3} & 63.9 & \textbf{67.1}                                                                       & \textbf{61.69}                                                                                 \\
                      \hline
\multirow{4}{*}{Plug} & CTC-CPPD & \multirow{4}{*}{SVTR-Base} & \multirow{4}{*}{$32\times100$}   & 61.3                 & 15.6                                                      & 53.2                 & 59.2                 & 69.4                 & 59.6                                                   & 63.8                 & 54.60                                             \\
                      & NRTR-CPPD &  & & 69.5                 & 21.0                                                      & 59.5                 & 64.2                 & 74.1                 & \textbf{65.5}                                                   & 67.0                 & 60.10                                             \\
                      & ABINet-CPPD &  & & 66.4                 & 19.1                                                      & 57.3                 & 56.7                 & 72.8                 & 55.8                                                   & 65.0                 & 56.19                                             \\
                      & PARSeq-CPPD &  & & 67.1                 & 19.0                                                      & 58.0                 & 58.0                 & 74.0                 & 61.9                                                   & 63.9                 & 57.19                                            
  \\

                      \hline
\end{tabular}
}

\label{tab:difficult_result}
\end{table*}

\noindent\textbf{The loss choice of CC and CO.}
As previously stated, a binary CE loss (i.e., Eq.~\ref{equ:bce}) is considered in the CO module. However, the CE loss described in Eq.~\ref{equ:ce2} also can implement this goal. We carry out an experiment to compare the two. As seen in Tab.~\ref{tab:loss}, the standard CE loss attains an accuracy of 93.50\%, 0.30\% lower than using our CO loss. The CO loss only concerns whether there are characters on the positions, while Eq.~\ref{equ:ce2} aims to identify the character at each position, making it more closely aligned with the final recognition objective. We argue that such an arrangement introduces the follow-up learning objective too early, and makes the feature enhancement somewhat vague. This result in turn justifies the rationality of our design.

On the other hand, ACE loss \cite{xie2019aggregation} can also achieve the goal of CC. The result in Tab.~\ref{tab:loss} shows that our CC loss gets a higher accuracy. As previously mentioned, this is mainly because the CC loss includes the appearance of all characters and is irrelevant to the character sequence length.

\noindent\textbf{The effectiveness of CC and CO.}
As shown in Tab.~\ref{tab:co}, nine experiments are designed to enumerate the combinations of CC/CO modules or/and losses. We take the model without both CC and CO modules as the baseline. As can be seen, when the two side losses are not employed, adding the CC or CO module individually also leads to an improvement of 0.48\% or 0.96\%, respectively, and applying both increases the accuracy by 1.04\%. This is attributed to the cross-attention-based learning, where a counting and/or ordering-based initialization also benefits the recognition to some extent. When the two side losses are enabled further, the CC or CO loss solely gives 0.86\% or 0.32\% additional gains, while they together report a prominent improvement of 1.79\% compared to the baseline. Besides, other enumerated combinations also report consistent improvements. The results demonstrate that the CC and CO modules are complementary and both take effect in CPPD. Specifically, the CC module targets character content, and the CO module emphasizes content-free reading order and positions. Combining them allows CC and CO to mutually reinforce and forms a more robust context description.

We further conduct experiments to analyze whether the CC and CO modules contribute to better feature extraction as follows. First, the complete CPPD is trained. Then, three variants are designed by appending a CTC decoder behind either $\mathbf{F}_{cc}$, $\mathbf{F}_{co}$ or $\mathbf{F}_{v}$, and discarding the rest parts of CPPD. Next, the three variants are trained under the constraint of freezing the entire network except for the CTC decoder. When converged, their performance on the six benchmarks is evaluated, as shown in Tab.~\ref{tab:ccfeatures}. Comparing to $\mathbf{F}_{v}$, decoding based on $\mathbf{F}_{cc}$ and $\mathbf{F}_{co}$ lead to 3.66\% and 1.9\% average accuracy gains, respectively. 

Since a decoder would biased toward the feature closer to the character prediction layer, we also devise another variant, termed as $\mathbf{\hat{F}}_{v}$, for a more in-depth comparison, where the CO embedding is used as \emph{query}, $\mathbf{F}_{v}$ as \emph{key} and \emph{value}, followed by the character recognition task. Therefore $\mathbf{\hat{F}}_{v}$ is closer than the $\mathbf{F}_{cc}$ and $\mathbf{F}_{co}$ above. However, it reports an accuracy of 89.93\%, worse than the results from the two features. The results clearly suggest that critical recognition context is modeled by CC and CO modules. 

\noindent\textbf{CC and CO visualization verification.}
We visualize attention maps of the final recognition features in Fig.~\ref{fig:cc_co_vis1}, where each of the 12 heads in MHSA is plotted from top to bottom. As can be seen, Head 1 (H1 for short) plays the role of character positioning and H0 mainly focuses on localizing the right-side character. For CPPD the hotspots are more accurate and obvious compared to those of CPPD without either CC or CO. In CPPD from H6 to H8 we see that many maps are responsible for splitting characters, which is much less observed in maps of the other three methods. This partially explains that CO and CC modules jointly build a more complete context and give better recognition. On the other hand, for AR-based NRTR, a majority of maps focus on character localization, which is in line with its character-by-character decoding nature.  

\subsection{Adaptability to Different Encoders}
We conduct three sets of experiments to validate the adaptability of CPPD to different encoders. For fair comparison, the experiments is carried out by appending CPPD to the encoder of popular STR models. First, CNN-based comparison. The same as ABINet \cite{fang2021abinet}, a ResNet45 and three Transformer units are taken as the encoder. As seen in Tab.~\ref{tab:encoder3}, using CPPD receives a accuracy gain of 1.5\%. Second, ViT-based comparison. The same ViT as PARSeq \cite{BautistaA22PARSeq} is employed. CPPD gives an improvement of 0.64\% in this case. Third, SVTR-B with different decoders. It is seen that both ABINet and PARSeq get better accuracy when SVTR-B is equipped as their encoder. Nevertheless, CPPD still gets the highest accuracy among the competitors. The results demonstrate that CPPD is well-suited to different encoders.

\subsection{SVTR-CPPD Series and Plugged Models}
Based on the analyses above, we construct two series of CPPD models for further experiments. First, the SVTR-CPPD series. SVTR-T-CPPD and SVTR-B-CPPD are constructed, which employ SVTR with different capacities as the encoder. They show different accuracy-speed balances and thus offer flexible choices for applications. Second, we integrate the proposed CC and CO modules into four popular STR decoders, i.e., CTC, NRTR \cite{Sheng2019nrtr}, ABINet \cite{fang2021abinet}, and PARSeq \cite{BautistaA22PARSeq}. Specifically, we employ SVTR-B as the encoder and use these decoders to replace the \emph{Transformer Block with MHA} in Fig.~\ref{fig:structrue}. To keep consistent with CPPD, $\mathbf{F}_{cc}$ is recruited as \emph{value} and \emph{key}, while the \emph{query} is also $\mathbf{F}_{cc}$ for CTC, the character embedding for NRTR, the fixed position embedding for ABINet and PARSeq. This series of models can validate the plug-and-play nature of CPPD modules.

\subsection{Comparisons on six English benchmarks}

\begin{figure*}[t]  
\centering  
\includegraphics[width=0.96\textwidth]{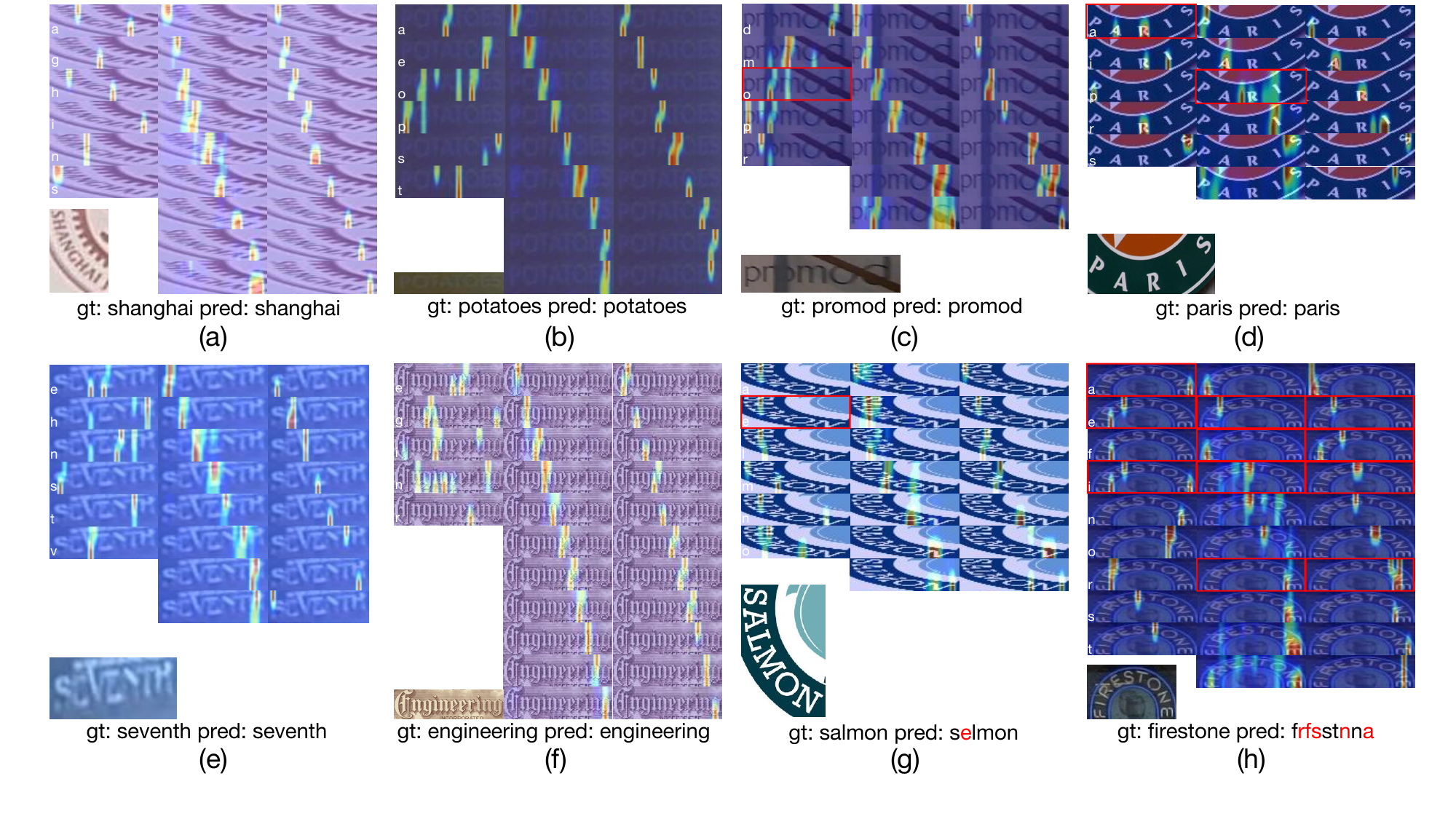}  
\caption{Visualization of attention maps. In each example, a text instance and its recognition result obtained from SVTR-B-CPPD are given. From left to right the three columns are visualization of CC, CO and the final recognition features, respectively. Among them, CC maps are vertically arranged in an alphabetical order, CO and recognition maps are arranged according to their reading orders. Red boxes denote the erroneously localized maps. The wrongly recognized characters are shown in red. The same below.}  
\label{fig:vis_2}  
\end{figure*}

We compare the constructed CPPD models with previous works on the six standard English benchmarks covering regular and irregular text instances in Tab.~\ref{tab:sota}. The previous works are grouped into three groups, i.e., AR, PD and CTC, for better analysis. While the two series of constructed CPPD models are categorized into "Ours" and "Plug". We first look into the SVTR-CPPD ones. There are 2.26\% and 2.2\% average accuracy gains from SVTR-T-CPPD v.s. SVTR-T, and SVTR-B-CPPD v.s. SVTR-B, respectively. The results clearly verify the effectiveness of CPPD. Note that the two SVTR-CPPD models do not incorporate the rectification module while SVTR did, which bring additional accuracy gains in general. Even larger accuracy gains may obtain when getting rid of the rectification module. Moreover, incorporating CPPD only introduces 2.26 M and 2.2 M increase in model size. Since the rectification module only has 1.88 M additional parameters, it implies that CPPD is a lightweight decoder and the introduced model size overhead is limited. When comparing SVTR-based AR and PD models, although SVTR-B-NRTR achieves a prominent average accuracy of 93.54\%, its CPPD counterpart, i.e., SVTR-B-CPPD, further gives 0.26\% accuracy gains, along with 7.75x inference acceleration and slightly smaller model size. Not only reaching the new state-of-the-art accuracy, but also running the fastest among the non-SVTR-based AR and PD models in Tab.~\ref{tab:sota}. Meanwhile, the accuracy gain observed by SVTR-T-CPPD v.s. SVTR-T-NRTR is 0.76\% with an 8.77x inference speedup. These accuracy advantages can be explained as AR decoding only models the context of already decoded characters while CPPD models all the characters. It thus builds a more comprehensive recognition context. In addition, SVTR-B-CPPD shows 0.32\% accuracy improvements and with smaller size and faster inference speed when compared to the recent LPV-B \cite{zhang2023linguistic}. The results again demonstrate the superiority of the CPPD architecture.

We then inspect the plugged models. CTC-CPPD surpasses SVTR-B by 1.12\% in terms of accuracy, implying the effectiveness of incorporating the CC and CO modules. The other three models also get improvement and achieve remarkable accuracy. For example, NRTR-CPPD outperforms SVTR-B-NRTR by 0.96\% and gets an average accuracy of 94.50\%, the highest so far. The result indicates that the CPPD context somewhat differs from the context of AR decoding, and they are still complementary to some extent. In addition, 2.08\% and 1.18\% accuracy gains are observed when comparing ABINet-CPPD with ABINet, and PARSeq-CPPD with PARSeq, respectively. By eliminating those caused by the encoder difference, there are still 0.99\% and 0.78\% gaps in these two comparisons (see Tab.~\ref{tab:encoder3}). The results indicate that the context generated by CPPD is not only effective, but also has great generalization capability and can be used by different decoders for further accuracy improvements.

\begin{table*}[t]\footnotesize
\caption{Results on Chinese text recognition (CTR) dataset tested against existing models, which come from \cite{yuICCV2023clipctr} except for the SVTR and CPPD series.}
\centering
\setlength{\tabcolsep}{5pt}{
\begin{tabular}{r|cccc|c|cc}
\hline
Method         & Scene & Web & Document & Handwriting & 
Average(\%) & Parameters($\times10^6$)  & FPS                             \\
\hline
CRNN \cite{shi2017crnn} (2017)           & 53.4  & 57.0 & 96.6 & 50.8 & 64.45   & 12.4 &                           \\
ASTER \cite{shi2019aster}   (2019)         & 61.3  & 51.7 & 96.2 & 37.0 &  61.55  & 27.2 &                           \\
MORAN \cite{pr2019MORAN}    (2019)        & 54.6  & 31.5 & 86.1 & 16.2  &  47.10  & 28.5 &                           \\
SAR \cite{li2019sar} (2019) & 59.7     & 58.0 & 95.7 & 36.5  & 62.48  & 27.8 &                          \\
SEED \cite{cvpr2020seed} (2020) & 44.7     & 28.1 & 91.4 & 21.0  &   46.30 & 36.1 &                          \\
MASTER \cite{pr2021MASTER}  (2021)           & 62.8 & 52.1 & 84.4 & 26.9 & 56.55 & 62.8 &                                \\
ABINet \cite{TPAMI2022ABINetPP}  (2021)           & 66.6 & 63.2 & 98.2 & 53.1 & 70.28  & 53.1 &                               \\
TransOCR \cite{cvpr2021TransOCR}   (2021)          & 71.3 & 64.8 & 97.1 & 53.0 & 71.55  & 83.9 &                               \\           
SVTR-B \cite{duijcai2022svtr} (2022)         & 71.7 & 73.8 & 98.2 & 52.2 & 73.98  & 26.3 &  175                             \\
SVTR-L \cite{duijcai2022svtr} (2022)        & 72.2 & 74.1 & 98.1 & 53.6 & 74.50  & 43.4 &   146                             \\
CCR-CLIP \cite{yuICCV2023clipctr}   (2023)         & 71.3 & 69.2 & 98.3 & \textbf{60.3} & 74.78  & 62.0 &   \\
\hline
SVTR-B-CPPD     & 74.4 & 76.1 & 98.6 &  55.3 & 76.10  & 29.4 & 188 \\
SVTR-B-CPPD-STN & \textbf{78.4}  & \textbf{79.3} & \textbf{98.9} & 57.6 & \textbf{78.55}  & 31.2 &   169   \\
\hline
\end{tabular}}

\label{tab:chinese}
\end{table*}

\subsection{Comparisons on Union14M-L benchmark}
We also conduct experiments on Union14M-L \cite{jiang2023revisiting}, a recent large-scale English benchmark that more realistically reveals the recognition challenges. The results trained based on synthetic datasets are given in Tab.~\ref{tab:difficult_result}. The accuracy of ABINet and PARSeq-B (using SVTR-B as the encoder), which perform well in the six standard benchmarks, is not high, revealing the challenges of large-scale realistic dataset. Compared with the results on the six benchmarks, different SVTR models exhibit large performance gaps, where 6.12\% and 5.38\% accuracy gains are obtained from SVTR-T to SVTR-B, and further to SVTR-L. A similar trend is also observed in the SVTR-NRTR and SVTR-CPPD series. The results suggest that a large-capacity visual encoder is more important on Union14M-L. The plugged models also report remarkable improvements, especially for PARSeq. It gives a prominent 4.57\% accuracy gain when CPPD modules are equipped, again demonstrating the effectiveness of CPPD. In addition, we observe that the resizing is a critical factor, where a large input resolution significantly improves the accuracy. For example, SVTR-B-CPPD improves the accuracy by 5.06\% when the resolution is switched from $32\times100$ to $48\times160$, highlighting a high-resolution input is a priority in challenging and accuracy-first scenarios.

\subsection{Comparisons on Chinese benchmark}
We then assess the models on the CTR benchmark. Ref.~\cite{yuICCV2023clipctr} gave the accuracy of nine existing methods as listed in Tab.~\ref{tab:chinese}, while the remaining SVTR and CPPD series are computed by ourselves. Note that the model size and inference time will increase a little for Chinese for the same method, due to involving a larger prediction layer and a large input resolution. However, both are not compromised by the vocabulary size due to the CC module not participating in the inference. SVTR-B-CPPD also shows accuracy gain. It gives a 2.12\% average accuracy gain compared to SVTR-B, while running faster due to not incorporating the rectification module. When STN is further considered, a prominent accuracy improvement of 4.57\% is obtained. This is because the CTR dataset exhibits strong disturbances including curvature, occlusion, light-dark contrast, etc. Merely adding an STN could contribute to an improvement of 2.45\% in terms of accuracy. Note that the two CPPD models outperform the recent CCR-CLIP, which is dedicated to CTR, by 1.48\% and 3.93\% in terms of accuracy. The results convincingly validate the great cross-language generalization ability of CPPD.

\subsection{Visualization Analysis}

\begin{figure}[t]  
\centering  
\includegraphics[width=0.48\textwidth]{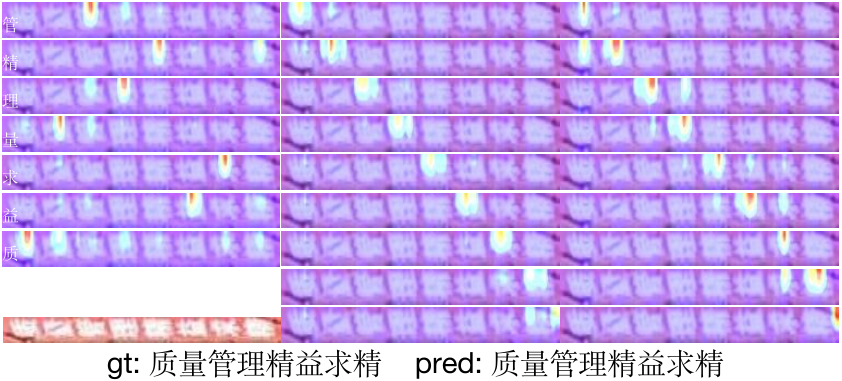}  
\caption{Visualization of Attention maps for a vertical Chinese text.}  
\label{fig:ch1}  
\end{figure}

\begin{figure}[t]  
\centering  
\includegraphics[width=0.48\textwidth]{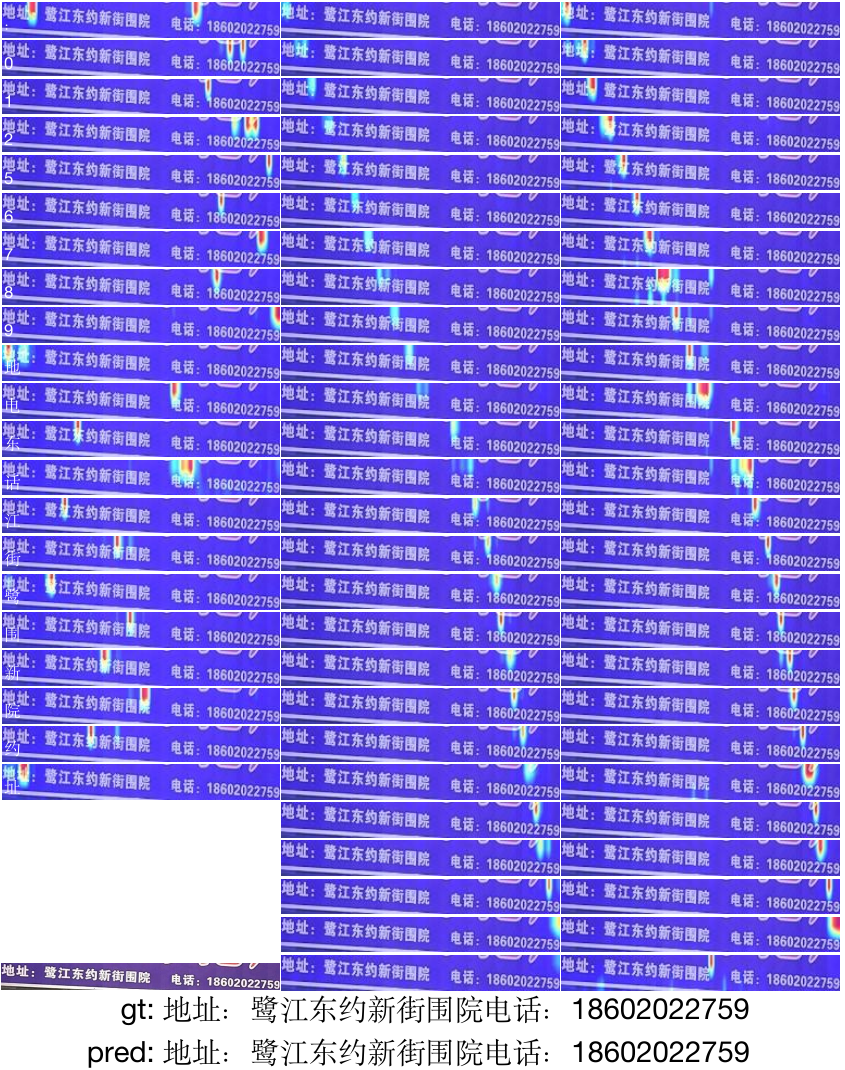}  
\caption{Visualization of attention maps for a long Chinese text. }  
\label{fig:ch2}  
\end{figure}

We give recognition results and attention maps of different CPPD components in Fig.~\ref{fig:vis_2} to verify the effectiveness of CPPD and whether its modules accomplish their task as desired. Specifically, eight challenging text instances are presented. The first six show the merit of CPPD. They exhibit excellent adaptation in severely curved, highly blurred, partially occluded, artistic, and large character space scenarios. Their attention maps show that the CC, CO, and recognition branches basically read the characters as expected. The first example, i.e., \emph{shanghai}, simultaneously exhibits text rotation, curvation, and blur, while our uniform resizing scheme introduces additional text deformation. Nevertheless, CPPD still correctly read the text. In the third and fourth examples, it is observed that mistakes are observed from a few CO or CC maps (marked by red boxes), e.g., the second \emph{o} is missed in the third CC map of Fig.~\ref{fig:vis_2}(c). Nevertheless, both cases are correctly recognized. The two examples illustrate the complementary nature of the three kinds of maps. Different features can complement each other and, to some extent, ward off distractions. There are also a few bad cases. For example the seventh example, after resizing, the image is severely deformed and character \emph{a} looks similar to \emph{e}, which causes the corresponding localization error in the CC map and mis-recognition. On the other hand, for the eighth example, which mixes several recognition difficulties such as severe curvation, fine character stroke, etc, hotspots of a portion of maps are mistakenly localized. For example, \emph{i} and \emph{e} are both counted as \emph{i} (and also \emph{e}), and the placeholders are not always presented in a left-to-right order. These factors lead CPPD to misidentify the word. We argue that strengthening the visual feature would be a way to correct these instances.

In Fig.~\ref{fig:ch1} and Fig.~\ref{fig:ch2}, we also give attention maps of two challenging Chinese text instances. The first is a blurred and vertically arranged Chinese text, which poses a great challenge to traditional CTR models. Nevertheless, all the maps give the right localization such that SVTR-B-CPPD correctly recognizes it. The other instance is a long Chinese text, whose recognition easily suffers from the attention drift \cite{yue2020robustscanner} in AR-based decoders and PD decoders also perform worse in general. When CPPD equipped, the hotspots in most maps correctly identify the characters and their placeholders, thus also giving a correct response. Both cases clearly validate how both CC and CO help the recognition. 

\section{Conclusion}
Aiming at developing accurate and fast STR models, we first analyze the differences between AR and PD decoders, from which we have inferred that character appearance and positions are valuable context variables for accurate recognition. Consequently, we have presented CPPD, a novel PD decoder that elegantly perceives these context variables using dedicated designed CC and CO modules and losses. Extensive experiments on English and Chinese text recognition convincingly validate our proposal. The constructed SVTR-CPPD models show remarkable balance on accuracy and inference speed, which not only perform top-tier in terms of accuracy, but also run approximately 8x faster than their AR-based counterparts. Moreover, the plugged models consistently give further accuracy improvement, especially on large and challenging datasets. Both demonstrate the effectiveness of the proposed CPPD.

While impressive experimental results are obtained from typical benchmarks, it is observed that the constructed context is still somewhat different and complementary from that built by AR decoders. Therefore, one future research objective is to further explore the missed context variables based on the CPPD framework. In addition, we are also interested in combining CPPD with recent large language or multi-modality models \cite{awais2023foundational, radford2021learning}, which also exhibit impressive understanding ability in OCR-related tasks. Integrating them together would be a topic worthy further study.

\section*{Acknowledgments}
This work was supported by the National Key R\&D Program of China (2022YFB3104703), and the National Natural Science Foundation of China (62172103).

\bibliographystyle{IEEEtran}
\bibliography{egbib}

\newpage

\vfill

\end{document}